%% file: main.tex
\documentclass[10pt,twocolumn,letterpaper]{article}

\usepackage{iccv}              


\definecolor{iccvblue}{rgb}{0.21,0.49,0.74}
\usepackage[pagebackref,breaklinks,colorlinks,allcolors=iccvblue]{hyperref}

\def\paperID{4544} 
\def\confName{ICCV}
\def\confYear{2025}

\title{NeuroADDA: Active Discriminative Domain Adaptation in Connectomics}

\author{
Shashata Sawmya\textsuperscript{1}
\quad Thomas L. Athey\textsuperscript{1}
\quad Gwyneth Liu\textsuperscript{1}
\quad Nir N Shavit\textsuperscript{1,2}\\
\textsuperscript{1}Massachusetts Institute of Technology
\quad \textsuperscript{2}Red Hat\\
{\tt\small \{shashata,\,tathey\_1,\,gwyliu,shanir\}@mit.edu}
}

\begin{document}
\maketitle
\input{sec/0_abstract}    
\input{sec/1_intro}

\input{sec/2_relatedWork_and_Experimental_setup}

\input{sec/3_Results}
\input{sec/4_discussion}

\input{sec/5_conclusion}
{
    \small
    \bibliographystyle{ieeenat_fullname}
    \bibliography{main}
}


\end{document}

%% file: sec/0_abstract.tex
\begin{abstract}

Training segmentation models from scratch has been the standard approach for new electron microscopy connectomics datasets. However, leveraging pretrained models from existing datasets could improve efficiency and performance in constrained annotation budget. In this study, we investigate domain adaptation in connectomics by analyzing six major datasets spanning different organisms. We show that, Maximum Mean Discrepancy (MMD) between neuron image distributions serves as a reliable indicator of transferability, and identifies the optimal source domain for transfer learning. Building on this, we introduce NeuroADDA, a method that combines optimal domain selection with source-free active learning to effectively adapt pretrained backbones to a new dataset. NeuroADDA consistently outperforms training from scratch across diverse datasets and fine-tuning sample sizes, with the largest gain observed at $n=4$ samples with a 25-67\% reduction in Variation of Information. Finally, we show that our analysis of distributional differences among neuron images from multiple species in a learned feature space reveals that these domain ``distances” correlate with phylogenetic distance among those species.
\end{abstract}

%% file: sec/1_intro.tex
\section{Introduction}
\label{sec:intro}

Connectomics is a field devoted to mapping neural circuits in tissue by reconstructing the ``wiring diagram” of the brain from high-resolution imaging data. In practice, this often involves imaging brain tissue volumes with electron microscopy (EM) to capture every neuron and synaptic connection at nanometer resolution. A critical step in this process is neuronal segmentation – partitioning the EM volume so that each voxel is assigned to the correct neuron or cellular structure \cite{Plaza2018}. Without reliable segmentation, reconstructing complete neural circuits at scale would be infeasible. 

Modern EM connectomics datasets are extremely large, often reaching hundreds of terabytes or even petabyte scales for just millimeters of brain tissue. Manual annotation of such massive image volumes is practically impossible – even a team of experts would require decades to delineate every neurite by hand \cite{Casser2020}. Although deep learning-based methods now produce automated segmentations, these are imperfect and require extensive human proofreading and correction to fix merge and split errors. In fact, error correction (proofreading) has become a major bottleneck: it can consume over 95\% of the total time and cost in a connectomics project \cite{Januszewski2018FFN}.  This heavy annotation burden is exacerbated by the lack of transfer learning in current workflows—models are usually trained from scratch on each new dataset. To date, transfer learning has been underexplored in connectomics, and there are no standardized guidelines for how to leverage pretrained neural networks for a new EM dataset.

Automated neuron segmentation methods typically use CNNs to predict cell boundaries in EM images, followed by voxel clustering to assign neuron IDs \cite{pavarino2023membrain, Beier2017Multicut}. Advances like the flood-filling network (FFN), a recurrent 3D CNN, have significantly improved accuracy by iteratively growing segments from seed points \cite{Januszewski2018FFN}. However, these methods usually suffer from poor generalization across different datasets due to domain shifts in imaging protocols or tissues. Consequently, new connectomics projects often can't leverage existing models, highlighting the need for better domain adaptation methods to avoid redundant annotation and training efforts.

Transfer learning and domain adaptation are highly promising for connectomics because they offer a way to leverage knowledge from existing labeled data to reduce the need for new annotations. In other computer vision and medical imaging tasks, transfer learning has been shown to significantly boost segmentation performance when training data in the target domain is limited \cite{Ardalan2022}. 
There is early evidence that domain adaptation on EM data can work: for instance; Barancco et. al. showed unsupervised domain adaptation in segmenting mitochondria in a new domain (fly images) from mouse \cite{franco2022deep} and Chacón et. al showed the same for synapse and mitochondria in small EM volumes \cite{BermudezChacon2020}. However, neuron segmentation is a distinct task, directly tied to connectome reconstruction, and presents additional challenges beyond organelle segmentation. Unlike prior works, which assumed optimal source-to-target domain transfer, our study addresses a more complex transfer setting where there are multiple source domains and only the best one needs to be chosen.


In this paper, we present the first attempt to tackle connectomic segmentation through active transfer learning, aiming to reduce the field’s reliance on uninformed training-from-scratch. We propose a framework that actively selects and transfers knowledge from a relevant source domain to improve segmentation on a target domain. In particular, we investigate optimal domain selection for transfer: given multiple candidate pretrained models on different source EM datasets from different brain regions, species, or imaging conditions; we determine which source domain yields the best performance boost on a new target tissue. After optimal domain selection, we design an algorithm that uses few annotated samples from the target domain to effectively transfer knowledge from the source to the target and optimize instance segmentation performance. Finally, we provide a preliminary analysis of the distributional differences between neuron images from various datasets. Interestingly, we find that the “distance” between image domains in a learned feature space correlates with meaningful biological distinctions.

%% file: sec/2_relatedWork_and_Experimental_setup.tex
\section{Related Works}
\label{sec:background}


\textbf{Active and Transfer Learning}

\begin{figure*}[ht]
    \centering
    \includegraphics[width=\linewidth]{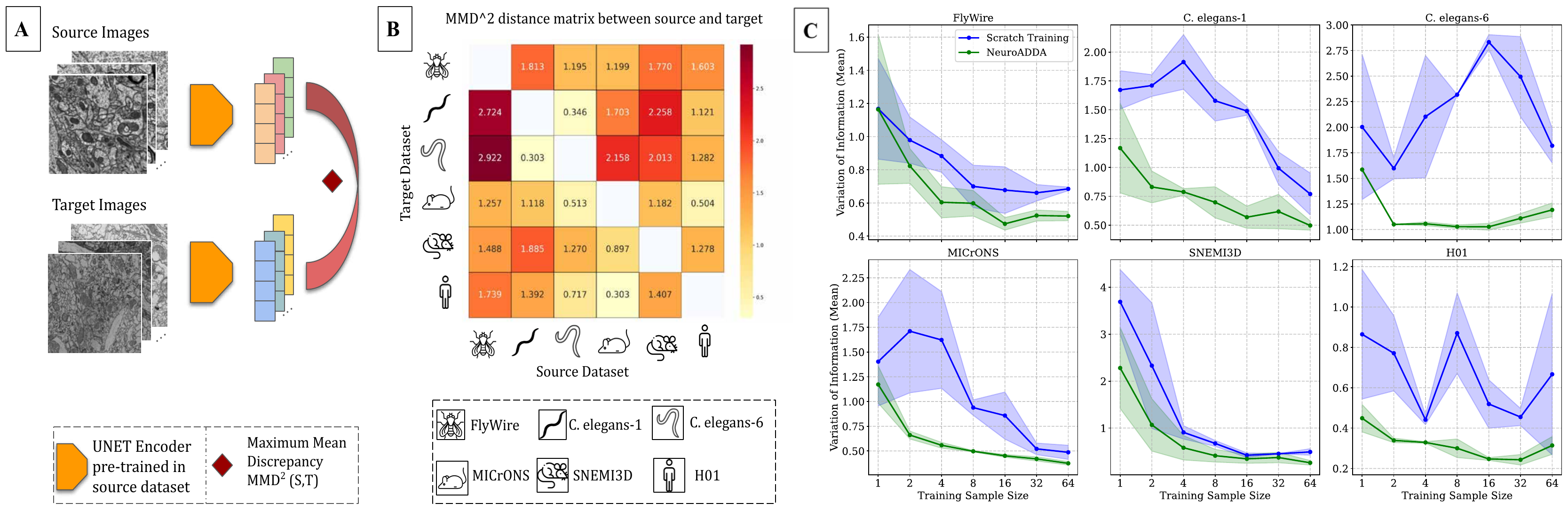}
    \caption{(A) Capturing the distribution shift between source and target images using source pretrained UNET Encoder. The distance is measured using squared Maximum Mean Discrepancy (B) The $MMD^2$ distance matrix illustrates the pairwise distribution shift between different datasets, where higher values indicate greater domain discrepancy. (C) Mean Variation of Information comparison (lower is better) between NeuroADDA (green) and scratch training (blue) across six datasets, showing that NeuroADDA consistently achieve better performance and lower variance over scratch training in lower training sample size.}
    \label{fig:ODS}
\end{figure*}

Active learning (AL) aims to maximize model performance while minimizing labeling effort by selecting the most informative unlabeled samples for annotation. Recent AL methods in computer vision combine uncertainty and diversity criteria to choose batches of points. For example, BADGE (Batch Active Learning by Diverse Gradient Embeddings) computes gradient embeddings for unlabeled samples and picks a batch that is both based on model uncertainty and diversity in this gradient space~\cite{Ash2020BADGE}. Similarly, CLUE (Clustering Uncertainty-Weighted Embeddings) clusters unlabeled samples in feature space while weighting by model uncertainty \cite{Prabhu2021CLUE}.

Given the scarcity of annotated data in many biological vision (biovision) tasks, AL is often combined with transfer learning (TL).
In practice, models pre-trained on large datasets (e.g. ImageNet) are fine-tuned on biomedical images to leverage learned features. Subsequently, AL can guide the labeling of the most informative new examples \cite{Ardalan2022}. Transfer learning is a crucial starting point because collecting and labeling huge biomedical image datasets is expensive \cite{Zhan2022AutoCSC}. There are instances of active transfer learning in other biovision taks. In a medical imaging classification setting, Hao et al. (2021) combined transfer learning with AL for brain tumor MRI grading, achieving slightly higher accuracy than a fully-supervised baseline while using 40\% fewer labeled examples \cite{Hao2021BrainTumorAL}. Roels and Saeys applied AL to 3D electron microscopy segmentation and reported 10–15\% higher Jaccard index than random sampling for the same labeling budget \cite{Roels2019ActiveEM}. Overall, active and transfer learning together form a powerful strategy to tackle the data scarcity in biological vision tasks, from cell classification to medical image segmentation, by reusing prior knowledge and judiciously choosing what to label next.

\textbf{Statistical Distance}
Many works in the literature compare distributions by analyzing deep activations via various metrics. For instance, Maximum Mean Discrepancy (MMD) has been used to bridge source and target distributions in adversarial training and domain adaptation \cite{gretton2012kernel}. Similarly, the Wasserstein distance (also referred to as optimal transport distance) is a popular measure for aligning intermediate embeddings, particularly in Wasserstein GANs, OT-based frameworks and even Large Language Models \cite{arjovsky2017wasserstein, courty2017joint, sawmyawasserstein}. Another widely adopted approach is CORrelation ALignment (CORAL), which matches the covariance statistics of source and target feature spaces for image distribution alignment \cite{sun2016return}. These metrics help quantify how effectively two distributions are brought into concordance, guiding both domain adaptation and representation learning in visual recognition tasks.

\textbf{Connectomic Segmentation}
Neuron segmentation from electron microscopy (EM) has rapidly advanced, surpassing human accuracy. Lee et al. \cite{lee2017superhuman}  achieved “superhuman” segmentation using a 3D U-Net with affinity-based agglomeration, while Beier et al. \cite{Beier2017Multicut} improved segmentation by combining local classifiers with global graph optimization. The Flood-Filling Network (FFN) by Januszewski et al. \cite{Januszewski2018FFN} refined segmentation continuity by iteratively growing neuron segments. In practice, FFN-based reconstructions produced much longer error-free neuron paths than earlier CNN+agglomeration pipelines, albeit with much higher computational cost. More recently, Sheridan et al. \cite{Sheridan2023LSD} introduced Local Shape Descriptors (LSDs), significantly enhancing accuracy while being $\sim$100× more efficient than FFNs. LSDs are a learned 10-dimensional representation capturing local geometry of neurons (e.g. process thickness, orientation) that serves as an auxiliary prediction task alongside boundary detection. These innovations push automated connectomics toward scalable, high-precision reconstructions of neural circuits. Despite these advances, most modern techniques, whether membrane segmentation, affinity prediction, or recurrent reconstruction-still rely on U-Net-based architectures. In this study, we adopt the membrane segmentation pathway, a widely used approach in this field.

In electron microscopy segmentation, various metrics have been explored, including the Variation of Information (VI) \cite{meirovitch2019cross, pavarino2023membrain}, Rand Index \cite{wang2022novel}, and Intersection over Union (IoU) \cite{Casser2020}. In this study, we focus on VI because it captures the instance-level merges and splits that arise from semantic membrane predictions, which directly affect connectomic segmentation quality. This makes VI particularly well-suited for measuring how well an automated segmentation aligns with the ground-truth neuronal instances in a large-scale connectomics setting.

\section{Experimental Setup}

We conducted experiments on six major connectomics datasets spanning four different organisms. Kasthuri et. al. (2015) \cite{kasthuri2015saturated} reconstructed a volume of the mouse somatosensory cortex, which was later released as the \textbf{SNEMI3D} dataset, making it one of the earliest publicly available connectomics datasets. Witvliet et al. (2021) 
\cite{witvliet2021connectomes} presented the reconstruction of a Nematode brain across different developmental stages, from which we selected two datasets: \textbf{C. elegans-1} (first larval stage, L1) and \textbf{C. elegans-6} (L3 stage). More recently, a complete reconstruction of the Drosophila melanogaster (fruit fly) brain, known as \textbf{FlyWire} \cite{schlegel2024whole}, was released, alongside \textbf{H01} \cite{shapson2024petavoxel}, a petabyte-scale reconstruction of the human cerebral cortex. Additionally, we incorporated data from \textbf{MICrONS} \cite{microns2021functional}, a fully reconstructed mouse visual cortex volume, which uniquely integrates both structural and functional information. Given the massive scale of these datasets—often comprising millions of images in the petabyte range—we selected non-overlapping subsets for our analysis, specifically from C. elegans-1, C. elegans-6, FlyWire, H01, and MICrONS. These subsets were chosen from regions with dense neuron segmentation. Our train and test sets are primarily drawn from consecutive slices. As SNEMI3D was released as a standalone dataset for segmentation model development in connectomics, we used its standard release \cite{lee2017superhuman}.

\cref{tab:datasets} presents details of the datasets used in this study, including species, tissue type, imaging modality, resolution, and the number of images. Notably, except for C. elegans-6, all datasets were imaged using Scanning Electron Microscopy (SEM); C. elegans-6 was the only dataset acquired with Transmission Electron Microscopy (TEM). Additionally, all samples have a uniform image size of (1024, 1024).

For segmentation, we use the fully convolutional U-Net architecture \cite{ronneberger2015u}, a widely adopted model for biomedical image segmentation. We focus on a 2D cell membrane segmentation pathway for EM images of brain tissue, where seeded watershed is applied to the predicted membrane maps to obtain instance segmentation and subsequently aid connectome reconstruction. However, in this study, our primary interest is the 2D membrane segmentation stage itself, for which we compare performance using the Variation of Information metric on the watershedded instance segmentation. Several U-Net models were trained on the aforementioned datasets under different training conditions (varying sample size and learning approach). All models were trained on a stack of 6 NVIDIA RTX 2080 GPUs.

\begin{table*}[ht]

\centering
\caption{Details of the datasets used in this study, including species, tissue type, imaging modality, resolution, and number of images.}
\resizebox{\textwidth}{!}{%
\begin{tabular}{lccccc}
\toprule
\textbf{Dataset} & \textbf{Organisms} & \textbf{Tissue} & \textbf{Imaging Modality} & \textbf{Resolution (X, Y, Z) [nm]} & \textbf{\# Images} \\ 
\midrule
SNEMI3D & Mus musculus & Somatosensory Cortex & SEM & (6.0, 6.0, 29.0) & 100 \\  
MICrONS & Mus musculus & Visual Cortex & SEM & (4.0, 4.0, 40.0) & 1335 \\  
H01 & Homo sapien & Cerebral Cortex & SEM & (4.0, 4.0, 33.0) & 2000 \\  
C. elegans-1 & C. elegans & Nerve Ring & SEM & (1.0, 1.0, 30.0) &  1603\\  
C. elegans-6 & C. elegans & Nerve Ring & TEM & (0.768, 0.768, 50.0) & 4708 \\  
FlyWire & D. melanogaster & Whole Brain & SEM & (4.0, 4.0, 40.0) & 1125 \\  
\bottomrule
\end{tabular}%
}
\label{tab:datasets}
\end{table*}

%% file: sec/3_Results.tex
\section{Results}

\subsection{Identifying Optimal Domain For Transfer}

To quantify the notion of similarity or dissimilarity between domains (datasets), consider a target domain \( T \) and a set of candidate source domains \( \{S_1, S_2, S_3, \dots, S_k\} \). Each source domain \( S_i \) has an associated pretrained segmentation model \( F(S_i) \), which serves as a representation of the domain. We then use \( F(S_i) \) as a feature extractor to embed images from both the source \( S_i \) and target \( T \). Specifically, we extract activations from the final down-projection layer of the U-Net and apply max-pooling to obtain a compact, fixed-length vector representation. This results in an embedding \( \phi(x) \) for each image \( x \in S_i \cup T \), which captures high-level semantic features of the neuron images.  As we have a vector representation for each datapoint in our source and target domain; we can now compute the difference between the image distributions. We used Maximum Mean Discrepancy (MMD), which is a kernel-based distribution discriminator distance metric. It quantifies the discrepancy between two probability distributions \( P \) and \( Q \) based on their embeddings in a Reproducing Kernel Hilbert Space (RKHS). A lower MMD value indicates a higher alignment between the source and target domains.  \cref{fig:ODS}A illustrates an example computation between two domains MICrONS and FlyWire.

Using this distance metric, we can identify the distribution shift for different organisms, and \cref{fig:ODS}B shows the distance matrix for all 6 datasets we explored in this study with lighter color indicating lower squared $\text{MMD}$ distance. For any target dataset, we hypothesize the optimal transfer domain as the one with the lowest $\text{MMD}^2$ value.

\subsection{Utilizing uncertainty quantification in sample selection: A free image artifact detector?}
After we identify the optimal source domain, we wanted to explore selecting informed samples that would be most suitable for finetuning in the target domain. Active learning literature \citep{Prabhu2021CLUE, Ash2020BADGE} explores uncertainty as a metric for selecting samples. To estimate uncertainty in a target dataset \( T \) using a pretrained source model \( F(S_i) \), we employ Monte Carlo (MC) Dropout, performing \( K \) stochastic forward passes for each input image \( x \) from the target domain. Given an input image \( x \), the model generates \( K \) probabilistic segmentation outputs:

\begin{equation}
P_k = \text{Softmax}(F(S_i, x, \theta_k)), \quad k = 1, 2, ..., K
\end{equation}

where \( F(S_i, x, \theta_k) \) represents the segmentation output of the source model \( F(S_i) \) under dropout-enabled parameters \( \theta_k \). The mean predictive distribution over the \( K \) runs is given by:

\begin{equation}
\bar{P} = \frac{1}{K} \sum_{k=1}^{K} P_k
\end{equation}

To quantify uncertainty, we compute the pixel-wise entropy:

\begin{equation}
H(x) = - \sum_{c=1}^{C} \bar{P}_c \log (\bar{P}_c + \epsilon)
\end{equation}

where \( C \) is the number of segmentation classes and \( \epsilon \) is a small constant for numerical stability. The average uncertainty across all pixels in an image is then given by:

\begin{equation}
U = \frac{1}{N} \sum_{i=1}^{N} H(x_i)
\label{eq:1}
\end{equation}

where \( N \) is the total number of pixels in the image. This entropy-based measure provides an estimate of segmentation uncertainty, with higher entropy indicating greater uncertainty in model predictions.

\begin{figure}
    \centering
    \includegraphics[width=\linewidth]{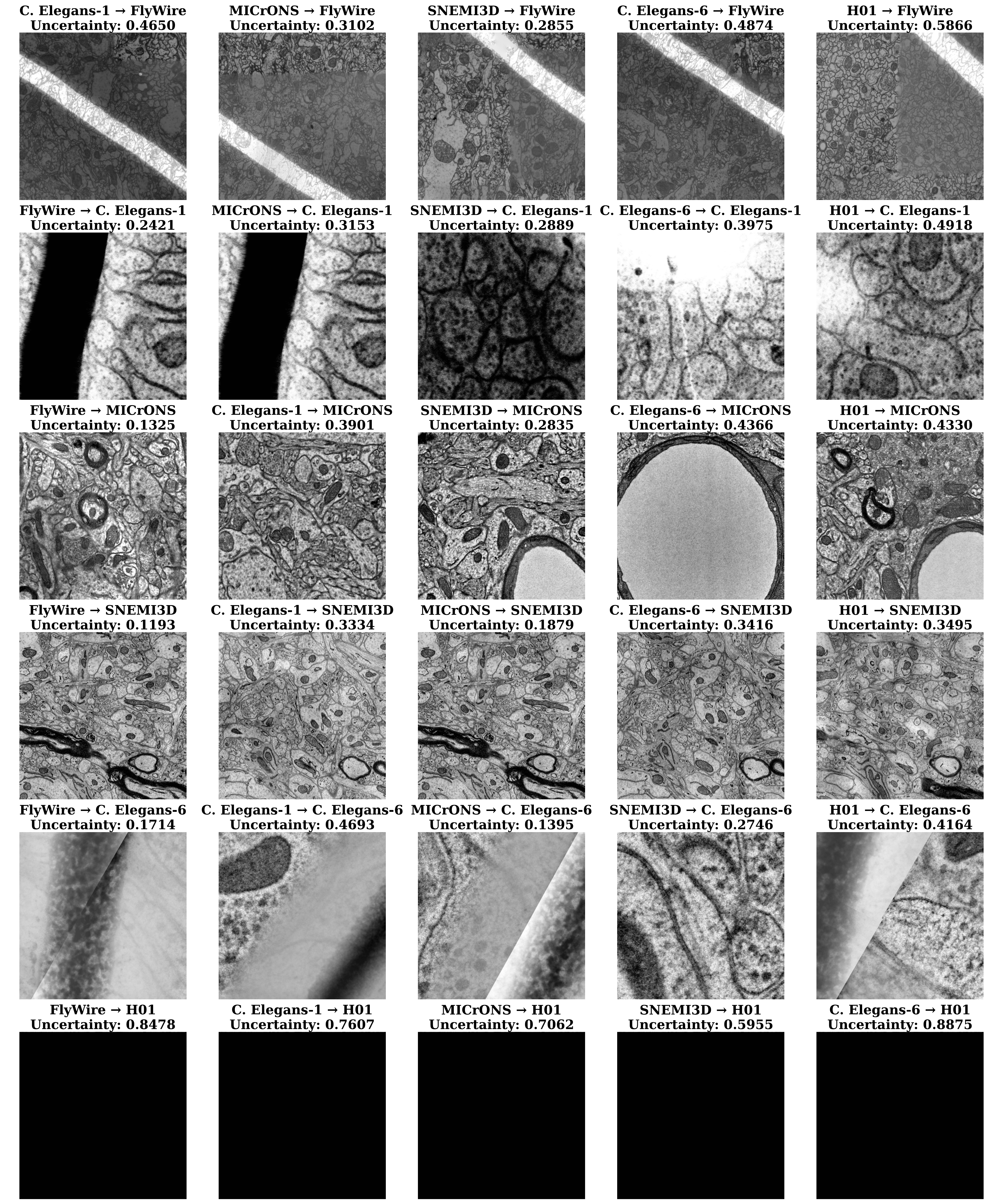}
    \caption{Examples of the most uncertain images under each source-to-target dataset transfer, accompanied by their corresponding uncertainty scores. In all transfer scenarios, high-uncertainty samples tend to be either image artifacts (e.g., severe brightness or contrast abnormalities) or biologically atypical regions (e.g., unusual structural patterns), indicating that such outliers pose the greatest challenge for model adaptation across datasets.}
    \label{fig:uncertain}
\end{figure}

We use the uncertainty metric in \cref{eq:1} to guide sample selection from the target domain while transferring from a selected source. Interestingly, the most uncertain images in a given source-to-target transfer exhibit a distinct pattern: they predominantly consist of image artifacts or particularly challenging samples within the target distribution. The most notable artifacts include white and black stripes running through the length of the image, contrast distortion, black tiles etc. (\cref{fig:uncertain}).

For target domains where samples are relatively clean (e.g., MICrONS and SNEMI3D), we still observe that the most uncertain images tend to have small distortions or contain more myelinated structures, which may contribute to segmentation difficulty. To systematically assess the nature of these artifacts, we manually categorized each image in the FlyWire dataset as major artifacts, minor artifacts, or clean samples. The annotator was blinded to the uncertainty scores. Our analysis confirms that high-uncertainty images overwhelmingly correspond to severe artifacts, most notably the bright stripes seen in FlyWire. To verify this, we performed a Mann–Whitney U test (\(\alpha=0.05\)) across four source domains, comparing the uncertainty scores of stripe‐afflicted images against those without stripes. The resulting \(p\)-values ranged from \(3.31\times10^{-10}\) to \(1.43\times10^{-32}\), denoting a \textbf{highly significant} difference. In other words, the presence of these white stripe artifacts consistently leads to higher model uncertainty in FlyWire, regardless of the particular source domain.

\subsection{Active Discriminative Domain Adaptation}
Building on our optimal domain selection framework and various sampling strategies, including uncertainty and diversity-based sampling, we propose an Active Learning algorithm dubbed \textit{NeuroADDA} (\textbf{Neuro}nal \textbf{A}ctive \textbf{D}iscriminative \textbf{D}omain \textbf{A}daptation) designed to achieve optimal performance in regimes where annotating data is expensive. The term discriminative refers to choosing the optimal domain from a set of candidate domains to transfer from.  In our experimental setting NeuroADDA performs better than scratch training or other passive or active transfer methods(see \cref{fig:ODS}C and \cref{tab:results_split}).

\cref{alg:neuroadda} introduces two additional constraints beyond standard active sample selection and iterative training: (1) the total number of annotated samples, denoted as \( A \), and (2) the training duration, denoted as \( B \). In our experiments, we evaluate different values of \( A \in \{1,2,4,8,16,32,64\} \), while keeping \( B = 10,000 \) gradient steps constant. We tested five sampling algorithms spanning both uncertainty-based and diversity-based strategies, and summarized their efficacy (the percentage of experimental settings in which each achieved the lowest mean VI) in \cref{tab:samplingAlgo}. Since median-uncertainty sampling was the top performer in most cases, we adopt it as our default strategy—except in passive learning, where we retain random sampling. As we tested multiple sampling algorithms within NeuroADDA, the general form of the algorithm can be referred to as NeuroADDA (\texttt{$\alpha$}), where \texttt{$\alpha$} represents any sampling strategy. Given that median-uncertainty achieved the best efficacy in our setup, we refer to NeuroADDA as the version using median-uncertainty sampling by default.


\begin{table}[ht]
\centering
\caption{Comparison of different sampling algorithms and their frequency of achieving the best performance.}
\resizebox{\linewidth}{!}{%
\begin{tabular}{lcc}
\toprule
\textbf{Sampling Algorithm} & \textbf{Strategy} & \textbf{Efficacy} \\ 
\midrule
Min-Uncertainty  & Uncertainty & 1.6\% \\  
Max-Uncertainty  & Uncertainty & 15.9\% \\   
CLUE  & Uncertainty + Diversity & 15.9\% \\  
BADGE  & Uncertainty + Diversity & 25.4\% \\  
Median-Uncertainty  & Uncertainty & \textbf{41.3\%} \\
\bottomrule
\end{tabular}%
}
\label{tab:samplingAlgo}
\end{table}


\begin{algorithm}[ht]
\caption{{\textbf{NeuroADDA} \textbf{A}ctive \textbf{D}iscriminative \textbf{D}omain \textbf{A}daptation}}
\label{alg:neuroadda}
\begin{algorithmic}[1]

\REQUIRE {Candidate pretrained models $\{M_1,\dots,M_N\}$ from $N$ domains},  unlabeled pool $U$, labeled pool $L$, number of \textit{active} iterations $T$,
         {annotation budget $A$, training budget $B$},
         sampling method $\texttt{SampleAlgo}$,
         Optimal Domain Selection routine $\texttt{ODS}$

\STATE {$M \gets \texttt{ODS}(\{M_1, \dots, M_N\}, U)$} // Choose best pre-trained model
\STATE {$T \gets \min(T, A)$} // Number of active iterations is at most annotation budget

\STATE {$L \gets \varnothing$ }  // Source-free

\STATE {$K \gets \frac{A}{T}$} // Annotations per iteration
      
\STATE {$\mathrm{S} \gets \frac{B}{T}$} // Training steps per iteration

\FOR{$t = 1$ to $T$}
  \STATE \textbf{1) Acquire unlabeled samples}
  \STATE \hspace{50pt} $S \gets \texttt{SampleAlgo}(M, U, K)$
  
  \STATE \textbf{2) Annotate}
  \STATE \hspace{50pt} Obtain ground-truth labels for $S$
  
  \STATE \textbf{3) Update labeled/unlabeled sets}
  \STATE \hspace{50pt} $U \gets U \setminus S$
  \STATE \hspace{50pt} $L \gets L \cup \{(x, \text{label}(x)) \mid x \in S\}$
  
  \STATE \textbf{4) Retrain or fine-tune the model for {$\mathrm{S}$} steps}
  \STATE \hspace{50pt} {Train model $M$ on labeled set $L$ for $\mathrm{S}$ gradient updates}

\ENDFOR

\RETURN $M$
\end{algorithmic}
\end{algorithm}

\subsection{Optimal domain adaptation outperforms scratch training}

\cref{tab:results_split} presents a comparative analysis of segmentation performance across six major connectomics datasets using four different training configurations (three for control):

\begin{enumerate}
    \item Scratch Training,
    \item Passive Transfer Learning with Minimum MMD (optimal domain),
    \item Active Transfer Learning with Maximum MMD (worst domain),
    \item Active Learning with Minimum MMD (optimal domain).
\end{enumerate}

For configurations (1) and (2), we employed random sampling, whereas for (3) and (4), we utilized median uncertainty sampling. The experiments were conducted across different training sample sizes ranging from \( 1 \) to \( 64 \). The results consistently demonstrate that scratch training performs worse than transfer learning across all sample sizes, highlighting the advantages of leveraging pretrained models for connectomics segmentation.

One particularly striking pattern in \cref{tab:results_split} is that the performance gap between scratch training and transfer learning methods is largest at smaller training sample sizes (S1–S16). This suggests that domain adaptation is especially beneficial when limited target data are available, where models trained from scratch struggle the most. As the sample size increases, this gap narrows somewhat, yet transfer learning consistently maintains an advantage across all conditions.

\begin{figure}[ht]
    \centering
    \includegraphics[width=\linewidth]{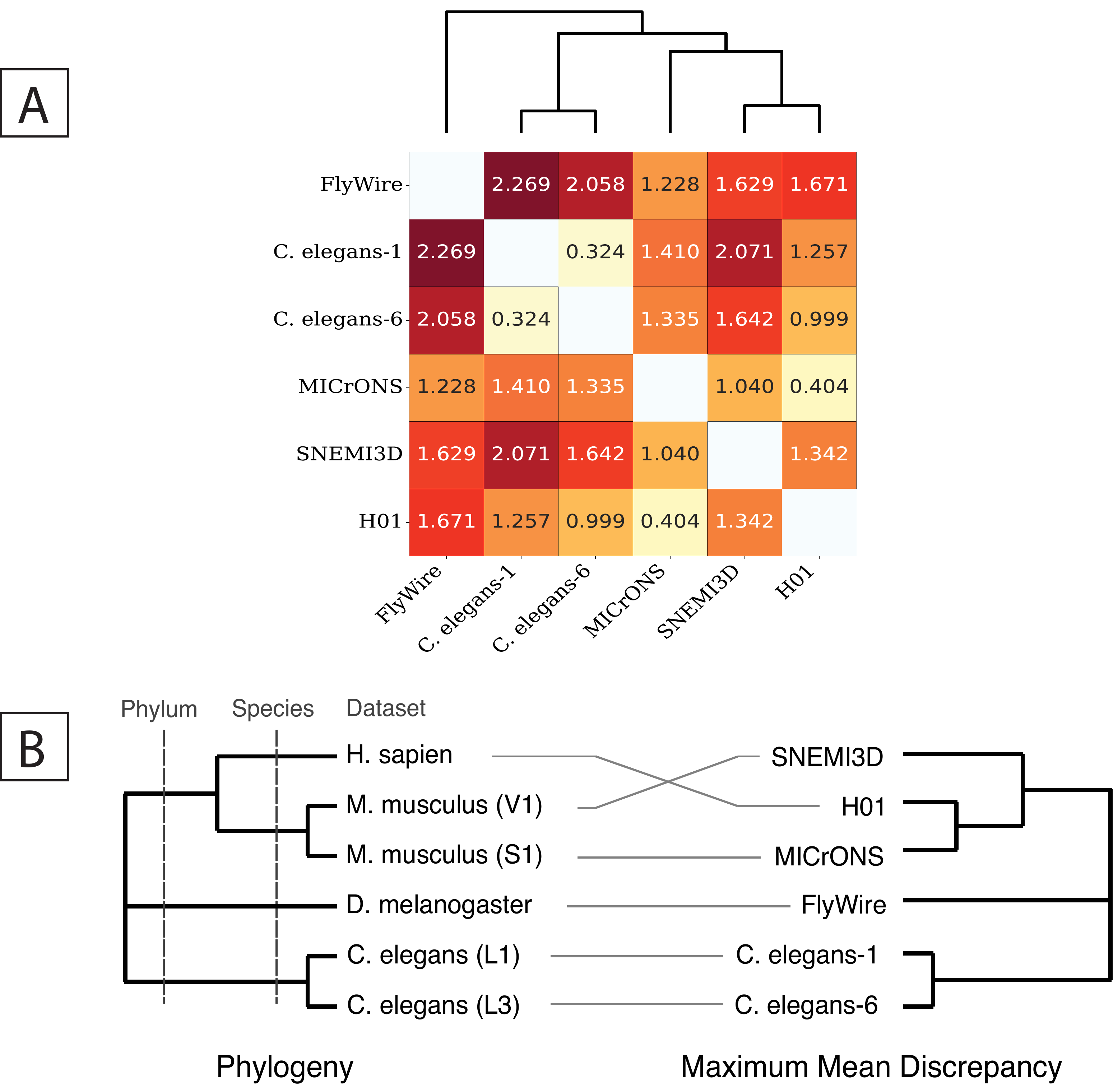}
    \caption{(A) Symmetrized MMD distance matrix and agglomerative clustering dendrogram across six major connectomics datasets, with darker squares representing greater domain distance. (B) Comparison between the resulting MMD‐based clustering and known phylogenetic relationships, highlighting that the two clusterings align significantly at the phylum level.}
    \label{fig:phylo_analysis}
\end{figure}

A key factor driving this improvement is optimizing the source domain for transfer, which proves to be highly effective regardless of whether active or passive learning is used. In fact, selecting the minimum MMD dataset yields the best performance in 99.2\% of cases, with active learning via median uncertainty sampling (NeuroADDA) providing additional gains in 83\% of those cases. This highlights the importance of both domain selection and sample efficiency strategies in achieving superior segmentation performance.

\begin{table*}[ht]
\centering
\resizebox{\textwidth}{!}{%
\begin{tabular}{cccccccccc}
\toprule
\textbf{Target} & \textbf{Learning Type} & \textbf{Transfer Domain} & \textbf{S1} & \textbf{S2} & \textbf{S4} & \textbf{S8} & \textbf{S16} & \textbf{S32} & \textbf{S64}
 \\
\midrule
FlyWire & Passive & Scratch Training & $1.168\pm0.304$ & $0.979\pm0.139$ & $0.883\pm0.098$ & $0.700\pm0.128$ & $0.677\pm0.141$ & $0.661\pm0.050$ & $0.685\pm0.011$ \\
FlyWire & Passive & min MMD & $1.358\pm0.064$ & $1.376\pm0.126$ & $0.936\pm0.022$ & $0.738\pm0.195$ & $0.570\pm0.007$ & $0.608\pm0.050$ & $0.628\pm0.073$ \\
FlyWire & Active & max MMD & $2.097\pm0.647$ & $1.032\pm0.124$ & $0.828\pm0.055$ & $0.805\pm0.161$ & $0.591\pm0.081$ & $0.543\pm0.047$ & $0.652\pm0.016$ \\
FlyWire & Active & min MMD & $\mathbf{1.164\pm0.451}$ & $\mathbf{0.823\pm0.105}$ & $\mathbf{0.604\pm0.094}$ & $\mathbf{0.598\pm0.077}$ & $\mathbf{0.474\pm0.038}$ & $\mathbf{0.524\pm0.033}$ & $\mathbf{0.521\pm0.028}$ \\
\midrule
C. elegans-1 & Passive & Scratch Training & $1.672\pm0.166$ & $1.710\pm0.094$ & $1.914\pm0.237$ & $1.578\pm0.176$ & $1.489\pm0.037$ & $0.993\pm0.141$ & $0.770\pm0.181$ \\
C. elegans-1 & Passive & min MMD & $\mathbf{0.897\pm0.038}$ & $1.096\pm0.359$ & $0.825\pm0.007$ & $0.791\pm0.022$ & $0.613\pm0.025$ & $\mathbf{0.365\pm0.023}$ & $\mathbf{0.370\pm0.014}$ \\
C. elegans-1 & Active & max MMD & $1.232\pm0.268$ & $1.268\pm0.178$ & $0.999\pm0.205$ & $0.882\pm0.127$ & $0.777\pm0.066$ & $0.729\pm0.100$ & $0.614\pm0.072$ \\
C. elegans-1 & Active & min MMD & $1.168\pm0.388$ & $\mathbf{0.831\pm0.138}$ & $\mathbf{0.788\pm0.029}$ & $\mathbf{0.698\pm0.135}$ & $\mathbf{0.569\pm0.096}$ & $0.618\pm0.149$ & $0.497\pm0.040$ \\
\midrule
C. elegans-6 & Passive & Scratch Training & $2.003\pm0.710$ & $1.598\pm0.103$ & $2.104\pm0.598$ & $2.318\pm0.000$ & $2.832\pm0.073$ & $2.494\pm0.392$ & $1.820\pm0.169$ \\
C. elegans-6 & Passive & min MMD & $\mathbf{1.155\pm0.019}$ & $1.356\pm0.084$ & $1.347\pm0.000$ & $1.169\pm0.040$ & $1.121\pm0.051$ & $1.109\pm0.062$ & $\mathbf{1.100\pm0.028}$ \\
C. elegans-6 & Active & max MMD & $1.873\pm0.231$ & $2.526\pm0.438$ & $1.707\pm0.140$ & $1.515\pm0.144$ & $1.567\pm0.069$ & $1.577\pm0.037$ & $1.274\pm0.078$ \\
C. elegans-6 & Active & min MMD & $1.586\pm0.004$ & $\mathbf{1.050\pm0.001}$ & $\mathbf{1.056\pm0.019}$ & $\mathbf{1.028\pm0.017}$ & $\mathbf{1.027\pm0.033}$ & $\mathbf{1.109\pm0.047}$ & $1.192\pm0.068$ \\
\midrule
MICrONS & Passive & Scratch Training & $1.403\pm0.448$ & $1.711\pm0.623$ & $1.622\pm0.489$ & $0.939\pm0.078$ & $0.858\pm0.238$ & $0.521\pm0.058$ & $0.487\pm0.071$ \\
MICrONS & Passive & min MMD & $1.110\pm0.382$ & $1.029\pm0.205$ & $0.863\pm0.002$ & $0.670\pm0.024$ & $0.565\pm0.088$ & $\mathbf{0.402\pm0.035}$ & $0.435\pm0.003$ \\
MICrONS & Active & max MMD & $0.918\pm0.112$ & $0.909\pm0.103$ & $0.640\pm0.040$ & $0.528\pm0.085$ & $0.512\pm0.044$ & $0.433\pm0.054$ & $0.416\pm0.048$ \\
MICrONS & Active & min MMD & $\mathbf{1.172\pm0.188}$ & $\mathbf{0.661\pm0.040}$ & $\mathbf{0.558\pm0.029}$ & $\mathbf{0.497\pm0.007}$ & $\mathbf{0.451\pm0.014}$ & $0.420\pm0.021$ & $\mathbf{0.375\pm0.017}$ \\
\midrule
SNEMI3D & Passive & Scratch Training & $3.690\pm0.688$ & $2.330\pm1.339$ & $0.908\pm0.142$ & $0.672\pm0.062$ & $0.420\pm0.049$ & $0.452\pm0.005$ & $0.491\pm0.069$ \\
SNEMI3D & Passive & min MMD & $3.075\pm0.828$ & $1.767\pm0.732$ & $0.821\pm0.161$ & $0.584\pm0.181$ & $0.491\pm0.112$ & $0.523\pm0.211$ & $0.334\pm0.009$ \\
SNEMI3D & Active & max MMD & $3.429\pm0.411$ & $3.255\pm0.520$ & $1.378\pm0.356$ & $0.568\pm0.088$ & $0.507\pm0.074$ & $\mathbf{0.322\pm0.025}$ & $0.333\pm0.074$ \\
SNEMI3D & Active & min MMD & $\mathbf{2.280\pm0.864}$ & $\mathbf{1.067\pm0.558}$ & $\mathbf{0.580\pm0.260}$ & $\mathbf{0.411\pm0.135}$ & $\mathbf{0.344\pm0.095}$ & $0.372\pm0.111$ & $\mathbf{0.260\pm0.048}$ \\
\midrule
H01 & Passive & Scratch Training & $0.865\pm0.321$ & $0.771\pm0.187$ & $0.442\pm0.024$ & $0.871\pm0.200$ & $0.519\pm0.120$ & $0.455\pm0.043$ & $0.667\pm0.399$ \\
H01 & Passive & min MMD & $\mathbf{0.351\pm0.030}$ & $0.401\pm0.018$ & $0.369\pm0.019$ & $0.673\pm0.300$ & $0.339\pm0.037$ & $0.244\pm0.016$ & $0.321\pm0.000$ \\
H01 & Active & max MMD & $0.706\pm0.213$ & $0.376\pm0.008$ & $0.555\pm0.082$ & $0.403\pm0.073$ & $0.312\pm0.055$ & $0.325\pm0.067$ & $0.333\pm0.083$ \\
H01 & Active & min MMD & $0.449\pm0.068$ & $\mathbf{0.339\pm0.011}$ & $\mathbf{0.329\pm0.004}$ & $\mathbf{0.300\pm0.046}$ & $\mathbf{0.247\pm0.007}$ & $\mathbf{0.243\pm0.026}$ & $\mathbf{0.314\pm0.044}$ \\
\bottomrule
\end{tabular}%
}
\caption{Comparison of transfer conditions for different sample sizes. Each cell shows the mean ± standard deviation. Bold entries denote
the minimum mean value for each target and sample-size column. $SX$ denotes $X$ samples used to train the model until convergence. Active Learning with min MMD transfer domain corresponds to our algorithm NeuroADDA.}
\label{tab:results_split}
\end{table*}

\subsection{MMD-based clustering and Phylogenetic Alignment}

In \cref{fig:phylo_analysis}A, we show the symmetrized MMD distance matrix for the six connectomics datasets (average of \cref{fig:ODS}B with its transpose), together with an agglomerative hierarchical dendrogram indicating how the species cluster based on learned feature embeddings. The color scale reflects domain distances (darker hues indicate greater dissimilarity), and we observe that the two C. elegans datasets cluster closely together even though they were imaged under different EM modalities. Mammalian datasets (M. musculus and H. sapiens) also form a coherent group, whereas D. melanogaster (FlyWire) stands farther apart. 

We performed a statistical test to compare the MMD hierarchical clustering with known phylogenetic relationships (\cref{fig:phylo_analysis}B). We considered two levels in hierarchy based on the number of clusters (\( k \)): the species clustering (\( k=4 \)), encompassing H. sapiens, M. musculus, D. melanogaster, and C. elegans, and the phylum clustering (\( k=3 \)), grouping these species in their respective phyla—chordates, arthropodes, and nematodes. Specifically, we used the clustering similarity criteria from Fowlkes and Mallows~\cite{fowlkes1983method}, and performed a permutation test to estimate the distribution of this statistic under the null hypothesis that the MMD clustering was independent of phylogeny.  

We permuted the cluster assignments while keeping the number of elements in each cluster fixed. Our test rejected the null hypothesis at \( k=3 \), indicating that MMD's perfect agreement with phylum assignments was statistically significant (\( p=0.019 \)). At \( k=4 \), the test failed to reject the null (\( p=0.261 \)).

%% file: sec/4_discussion.tex
\section{Discussion}
\label{sec:discuss}

Our experiments demonstrate clear benefits of transferring from an optimal source domain instead of training from scratch, especially when few labeled examples are available (\cref{tab:results_split}). Active Transfer Learning using minimum MMD (selecting the most similar source domain) consistently yielded higher segmentation accuracy and reduced error by 20–60\% relative to training from scratch, particularly with very low sample sizes (2, 4, 8, or 16 annotated images). This advantage persisted across different organisms (e.g., C. elegans, Drosophila, mammalian datasets), underscoring domain similarity as a strong predictor of transfer success.

Conversely, choosing a mismatched source domain (high MMD distance) led to poorer performance, sometimes worse than training from scratch at higher sample sizes. These findings emphasize the importance of quantifying domain distances rather than assuming that any pretrained model is sufficient.

\cref{fig:phylo_analysis} confirms that biologically related species cluster together in feature space, reflecting known phylogenetic relationships. For example, despite imaging differences (serial section TEM vs. SEM), C. elegans datasets clustered tightly, indicating that our feature-based domain distance captures fundamental neural characteristics rather than imaging modality differences. Similarly, mammalian datasets (mouse and human cortex) grouped closely together, distinct from nematode or fruitfly data. Computationally, this indicates that the U-Net backbone learned representations sensitive to biologically relevant domain shifts, capturing morphological traits like neuron size and tissue organization.

Despite these promising findings, our analysis is not without limitations. First, we could not fully disentangle the influence of imaging parameters (e.g. electron microscopy mode, section thickness, staining protocol) on the computed domain distances. While our results point to meaningful biological signals, a more controlled study is needed to explicitly quantify and mitigate any confounding effect of imaging settings. Second, our current analysis only includes four species, limiting the statistical power and generalizability of the observed domain clusters. Expanding to additional organisms and broader datasets would provide a more robust foundation for claims about cross-species distinctions. Finally, although we used a 2D U-Net backbone to compute features, other segmentation pathways (e.g. 3D recurrent CNNs, affinity-based prediction networks) could yield more \textit{informed} embeddings. Exploring these alternatives would help confirm whether the observed domain shifts reflect fundamental properties of neural tissue rather than methodological idiosyncrasies.


Though not without limitations, our findings carry significant implications. Domain-distance clustering aligned with phylogeny suggests embedding-based clustering could be a powerful tool for biological discovery, enabling comparisons across developmental stages or species—akin to ``image-based phylogenetics." Practically, researchers can use clustering to optimally select pretrained models by choosing sources within the same or nearest cluster, enhancing segmentation accuracy and reducing required labeling effort. For example, initializing from a similar pretrained model achieved comparable accuracy with only 4 labeled samples, which would otherwise require 16 or more if trained from scratch. Such efficiency gains could significantly reduce human labeling efforts and turnaround times for large connectome reconstructions. Finally, clustering provides a strategic blueprint for transfer learning, integrating machine learning metrics (domain distribution distances) with biological insights (evolutionary context) to optimize segmentation workflows in connectomics.

%% file: sec/5_conclusion.tex
\section{Conclusion}

In conclusion, this study introduced an active transfer learning framework for connectomics segmentation that leverages optimal domain selection. Our key contributions are two-fold: (1) we demonstrated that a simple metric – the Maximum Mean Discrepancy between high-level image features – can reliably predict the most transferable source domain for a new target dataset, and (2) we combined this insight with an active learning strategy to efficiently adapt segmentation models with minimal labeling effort. By systematically analyzing six diverse EM datasets (spanning nematode, insect, mouse, and human brains), we showed that using a well-chosen pretrained model from a similar domain consistently improves segmentation performance over training from scratch. Notably, our approach (referred to as NeuroADDA) achieved significant gains in the low annotation-data regime: for example, with only a handful of labeled samples (e.g. 4–8 annotations) it reduced segmentation error by 25–67\% compared to scratch training, a substantial improvement for practical connectomics workflows. These results underline the value of active domain adaptation in reducing the annotation burden in large-scale neural circuit reconstruction.


Looking forward, our approach opens up several avenues for future work. First, we envision developing more advanced embedding-based models to analyze tissue-wide and evolutionary differences across connectomic datasets. While our use of a U-Net’s feature activations proved effective in capturing domain relationships, future models could employ self-supervised learning or graph-based embeddings to encode even richer information about neural morphology. Such embeddings might allow researchers to quantitatively compare entire brains or substructures across species and developmental stages, potentially revealing new biological insights (for example, clustering various cortical regions or comparing healthy vs. diseased tissue in feature space). 
Second, experimental validation will be key to strengthening our conclusions. Applying our optimal domain selection and active adaptation pipeline to a new EM dataset could empirically confirm reductions in required annotations and improvements in segmentation quality. Additionally, examining learned feature representations for correlations with known histological differences, such as synapse density or myelination, would provide further biological validation. 

Lastly, we aim to refine the domain alignment techniques themselves as part of the adaptation process. In this work, we primarily used Maximum Mean Discrepancy (MMD) as a distance and focused on active fine-tuning of the source model on target data. Future approaches could integrate explicit domain adaptation methods (such as adversarial feature alignment or normalization statistic matching) to further close the gap between source and target domains before querying labels. By doing so, even when faced with a target domain that is somewhat distant, the model could automatically adjust its representations to become more target-like, thereby improving subsequent active learning efficiency. We also see potential in multi-source adaptation: if a new dataset lies between multiple known domains, techniques that can draw on multiple pretrained models (weighted by their similarity or through ensemble learning) might yield an even better starting point than any single source. In summary, our study demonstrates that active domain adaptation is a powerful paradigm for connectomics – one that exploits both the commonalities and differences between datasets. By uniting computational methods for optimal transfer with an appreciation of underlying biological relationships, we can significantly accelerate the mapping of connectomes and enhance our understanding of the brain’s wiring across different species and conditions.